\documentclass[runningheads]{llncs}
\usepackage[T1]{fontenc}
\usepackage{pdflscape}
\usepackage{amsmath}
\usepackage{graphicx}
\usepackage{amssymb}   

\begin{document}
\title{Graph-Guided Universum Learning in Generalized Eigenvalue Proximal SVMs for Alzheimer's Disease Classification}
\author{
Yogesh Kumar \and
Vrushank Ahire \and
Mudasir Ganaie\thanks{Corresponding author.}
}

\authorrunning{Y. Kumar et al.}

\institute{
Dept. of Computer Science and Engineering, IIT Ropar,
Punjab 140001, India\\
\email{
\{yogesh.23csz0014, 2022csb1002, mudasir\}@iitrpr.ac.in
}
}

\maketitle              
\begin{abstract}

Early and accurate detection of Alzheimer's disease (AD) is important for timely intervention and disease management. Generalized Eigenvalue Proximal Support Vector Machine (GEPSVM) and its Universum-based variants have shown promising results for AD classification. However, existing methods treat Universum samples as independent points and do not consider the geometric relationships among them. This paper proposes two graph-guided Universum learning models, namely UG-GEPSVM and IUG-GEPSVM, for AD versus cognitively normal (CN) classification using structural MRI data. In the proposed framework, mild cognitive impairment (MCI) subjects are used as Universum data to provide intermediate information between AD and CN classes. A graph is constructed over the Universum samples using Gaussian similarity, Minimum Spanning Tree connectivity, and multi-hop propagation. From this graph, a Laplacian matrix is derived that captures the geometric structure of the MCI samples. This Laplacian-based regularization is incorporated into the learning process in place of the conventional independent Universum penalty term. UG-GEPSVM integrates this regularization into the generalized eigenvalue formulation, while IUG-GEPSVM extends the numerically stable improved GEPSVM framework using a standard eigenvalue formulation. Experiments on ADNI MRI dataset variants using ICA- and PCA-based features at five different noise levels show that both proposed models consistently outperform existing GEPSVM and Universum-based methods. UG-GEPSVM achieves the highest average AUC of 88.07\% and maintains stable performance under increasing noise levels. Statistical tests further confirm the significance of the observed improvements. Code will be made publicly available upon acceptance.

\end{abstract}
\section{Introduction}
Alzheimer's disease (AD) is a progressive neurodegenerative disorder and one of the leading causes of dementia worldwide. According to the World Alzheimer's Report 2022, nearly 55 million people were living with dementia in 2019, and this number is expected to reach 139 million by 2050 \cite{alzheimer2022world}. Since AD remains incurable, early and accurate diagnosis plays an important role in timely intervention and disease management \cite{kumar2021alzheimer}. Mild cognitive impairment (MCI) is generally considered an intermediate stage between cognitively normal (CN) aging and AD, and provides useful prior information for automated AD versus CN classification systems. 

Structural MRI provides valuable biomarkers for AD classification. However, high-dimensional features, class imbalance, and limited labeled samples make learning challenging. Classical SVMs suffer from high computational cost and are restricted to parallel decision boundaries. To address these limitations, Mangasarian et al.~\cite{mangasarian2006multisurface} introduced the Generalized Eigenvalue Proximal Support Vector Machine (GEPSVM), which constructs two non-parallel proximal hyperplanes by solving two generalized eigenvalue problems. To improve numerical stability, Shao et al.~\cite{shao2012improved} proposed IGEPSVM, which replaces the ratio-based objective with a difference-based standard eigenvalue formulation and allows independent control of class-wise contributions.

To improve generalization when labeled data are limited, Universum learning incorporates prior information through samples that do not belong to either target class. Introduced by Weston et al.~\cite{weston2006inference} and further analyzed by Chapelle et al.~\cite{chapelle2007analysis}, the framework has been extended to twin hyperplane classifiers through U-TSVM~\cite{qi2012twin}. Several Universum-based variants, including fuzzy ULSTSVM~\cite{richhariya2022fuzzy}, intuitionistic fuzzy IFUTSVM-ID~\cite{tanveer2024robust}, and GBU-TSVM~\cite{ganaie2025granular}, have demonstrated improved generalization. In AD versus CN classification, MCI subjects naturally serve as Universum samples because they exhibit intermediate characteristics between the two classes. Richhariya et al.~\cite{richhariya2020diagnosis,richhariya2021efficient} showed that MCI-derived Universum samples improve feature selection and classification performance on ADNI datasets. More recently, Kumar et al.~\cite{kumar2026unified} proposed U-GEPSVM and IU-GEPSVM, which incorporate Universum constraints into the generalized eigenvalue framework and improve performance over GEPSVM and Universum-TSVM variants.

Despite these developments, an important limitation remains in existing Universum-based SVM methods, including U-GEPSVM and IU-GEPSVM. Universum samples are treated as independent and equally weighted points, without considering the geometric relationships among them. Graph-based learning and manifold regularization \cite{belkin2006manifold,kipf2016semi,wu2019simplifying} address this issue by modeling structural relationships through graph Laplacian penalties that encourage nearby samples to have similar representations. Recent studies have shown that MST-based graph construction combined with multi-hop aggregation effectively captures both local and global information from Universum samples \cite{richhariya2026gulstsvm}. Since MCI subjects often form meaningful neighborhoods in brain morphometric space, graph connectivity can provide additional structural information for Universum regularization. However, graph-guided Universum learning has not yet been explored within the generalized eigenvalue proximal SVM framework.

Motivated by these observations, this paper proposes two graph-guided Universum GEPSVM models: Universum Graph-Guided Generalized Eigenvalue Proximal Support Vector Machine (UG-GEPSVM) and Improved Universum Graph-Guided Generalized Eigenvalue Proximal Support Vector Machine (IUG-GEPSVM). Both models construct an MST-based graph over MCI Universum samples and employ multi-hop propagation to capture higher-order relationships. The resulting graph Laplacian is incorporated into the Universum regularization term, which allows the classifier to exploit both Universum information and its underlying geometric structure. UG-GEPSVM incorporates this graph-guided regularization into the ratio-based GEPSVM framework forming a generalized eigenvalue problem. IUG-GEPSVM extends the numerically stable IGEPSVM formulation and reduces learning to a standard eigenvalue problem.

The main contributions of this work are as follows:
\begin{enumerate}
\item UG-GEPSVM incorporates the graph-guided Universum regularization into GEPSVM for AD versus CN classification and formulates the problem as a generalized eigenvalue problem.
\item IUG-GEPSVM extends IGEPSVM with graph-guided Universum regularization while preserving its numerically stable standard eigenvalue formulation.
\item MST-based multi-hop graph Laplacian captures the geometric structure of MCI samples and their intermediate position between CN and AD.
\item Experiments and analyses on ADNI~\cite{mueller2005alzheimer} dataset demonstrate that the proposed models outperform existing methods, including under noisy conditions.
\end{enumerate}

The paper is organized as follows. Section~\ref{related_work} reviews related work, Section~\ref{methodology} presents the proposed models, Section~\ref{experiments} reports the experimental results and analysis, and Section~\ref{conclusion} concludes the paper.

\section{Related Work}
\label{related_work}

This section briefly reviews the Universum-based extensions of the GEPSVM and IGEPSVM classifiers, as well as graph-based regularization techniques that form the foundation of the proposed graph-guided Universum learning framework. Let $X_1 \in \mathbb{R}^{m_1 \times n}$ and $X_2 \in \mathbb{R}^{m_2 \times n}$
denote the positive and negative class data matrices, respectively,
$U \in \mathbb{R}^{p \times n}$ the Universum data matrix, and
$e_1 \in \mathbb{R}^{m_1}$, $e_2 \in \mathbb{R}^{m_2}$, $e_u \in \mathbb{R}^{p}$
vectors of ones.

\subsection{Universum Generalized Eigenvalue Proximal Support Vector Machine (UGEPSVM)}

UGEPSVM \cite{kumar2026unified} integrates Universum learning into the GEPSVM framework. UGEPSVM extends the GEPSVM formulation to include an additional Universum penalty term. For the first hyperplane, the optimization problem is:

\begin{equation}
\min_{(w_1,b_1) \neq 0}
\frac{\|X_1 w_1 + e_1 b_1\|^2 + \delta \|[w_1;\,b_1]\|^2}
{\|X_2 w_1 + e_2 b_1\|^2 + \|U w_1 + e_u b_1\|^2}
\end{equation}
Defining $G=[X_1\ e_1]^T[X_1\ e_1]+\delta I$, $H=[X_2\ e_2]^T[X_2\ e_2]$, and $P=[U\ e_u]^T[U\ e_u]$, the problem reduces to the generalized eigenvalue problem $Gz_1=\lambda_1(H+P)z_1$. The second hyperplane is obtained analogously via $Mz_2=\lambda_2(N+Q)z_2$, where $M=[X_2\ e_2]^T[X_2\ e_2]+\delta I$, $N=[X_1\ e_1]^T[X_1\ e_1]$, and $Q=[U\ e_u]^T[U\ e_u]$. In the AD versus CN setting, MCI subjects serve as Universum data, providing intermediate structural information between the two target classes.

\subsection{Improved Universum Generalized Eigenvalue Proximal Support Vector Machine (IUGEPSVM)}

IUGEPSVM \cite{kumar2026unified} extends IGEPSVM with Universum learning through a weighted difference-based formulation that provides independent control over class discrimination and Universum regularization. The objective function for the first hyperplane is:

\begin{equation}
\min_{(w_1,b_1) \neq 0}
\left(
\frac{\|X_1 w_1 + e_1 b_1\|^2}{\|w_1\|^2 + b_1^2}
-\gamma_1 \frac{\|X_2 w_1 + e_2 b_1\|^2}{\|w_1\|^2 + b_1^2}
-\psi_1 \frac{\|U w_1 + e_u b_1\|^2}{\|w_1\|^2 + b_1^2}
\right)
\end{equation}

where $\gamma_1 > 0$ controls class discrimination and $\psi_1 > 0$ controls the Universum contribution. Using the same matrix definitions as UGEPSVM and adding Tikhonov regularization, this reduces to the standard eigenvalue problem: $\left[(G+\delta I)-\gamma_1H-\psi_1P\right]z_1=\lambda_1 z_1$ and similarly $\left[(M+\delta I)-\gamma_2N-\psi_2Q\right]z_2=\lambda_2 z_2$ for the second hyperplane. The eigenvectors corresponding to the smallest eigenvalues give the optimal hyperplane parameters. Compared to UGEPSVM, IUGEPSVM provides improved numerical stability by solving standard rather than generalized eigenvalue problems, while the separate weighting parameters $\gamma_i$ and $\psi_i$ offer greater flexibility in balancing class separation and Universum alignment.

\subsection{Graph Laplacian Construction from Universum Data}
The graph Laplacian $L \in \mathbb{R}^{p \times p}$ is constructed from Universum data~\cite{richhariya2026gulstsvm}. The construction proceeds in three stages. First, a Gaussian weighted pairwise similarity matrix $D \in \mathbb{R}^{p \times p}$ is computed, where each entry is defined as
$D_{ij}=\exp\!\left(-\frac{\|U_i-U_j\|^2}{2\sigma^2}\right)$. Second, Prim's algorithm is applied to $D$ to obtain a Minimum Spanning Tree (MST), yielding a connected undirected graph with $p-1$ edges. Self-loops are added to form the adjacency matrix $\hat{A}=A_{\text{MST}}+I_p$, and the symmetrically normalized adjacency matrix is computed as $\hat{N}=\hat{D}^{-1/2}\hat{A}\hat{D}^{-1/2}$, where $\hat{D}$ is the degree matrix of $\hat{A}$. Third, multi-hop aggregation is applied for $h$ iterations using the update rule $A^{(t)}=\hat{N}A^{(t-1)}$, with $A^{(0)}=\hat{A}$, enabling propagation of structural information across the graph. The resulting matrix is symmetrized as $A'_{ij}=\max(A^{(h)}_{ij},A^{(h)}_{ji})$, and the graph Laplacian is computed as $L=D'-A'$, where $D'$ is the degree matrix of $A'$.

\section{Proposed Universum GEPSVM Models}
\label{methodology}
 
This section presents graph-guided extensions of U-GEPSVM and IU-GEPSVM that incorporate structural information derived from the Universum data into the generalized eigenvalue learning framework.  A graph Laplacian matrix
$L \in \mathbb{R}^{p \times p}$ is constructed from the Universum samples using MST connectivity and multi-hop aggregation.  Unlike conventional Universum-based methods that
treat each Universum sample in isolation, the proposed models exploit the
geometric relationships among these samples to guide hyperplane construction.
In the AD versus CN classification problem, MCI subjects serve as Universum
data and provide intermediate structural information between the two classes,
contributing to smoother decision boundaries and improved generalization.
 
\subsection{Universum Graph-Guided Generalized Eigenvalue Proximal Support
            Vector Machine (UG-GEPSVM)}
 
Standard UGEPSVM penalizes Universum samples through an independent
squared-distance term, ignoring the geometric relationships among them.
UG-GEPSVM addresses this limitation by replacing that term with a graph
Laplacian regularization $\|L^{1/2}(Uw_1 + e_u b_1)\|^2$. This penalizes
differences in the hyperplane projections of connected Universum samples and
thereby enforces smoothness along the Universum manifold.  The optimization
problem for the first hyperplane is:
 
\begin{equation}
\min_{(w_1,\,b_1)\neq 0}
  \frac{\|X_1 w_1 + e_1 b_1\|^2 + \delta\,\|[w_1;\,b_1]\|^2}
       {\|X_2 w_1 + e_2 b_1\|^2 + \|L^{1/2}(U w_1 + e_u b_1)\|^2}
\label{eq:ug_obj1}
\end{equation}
 where $\delta > 0$ is a Tikhonov regularization parameter and $L$ is symmetric
positive semidefinite.  Expanding the graph-regularized denominator term using
the properties of the graph Laplacian gives:
 
\begin{align}
\|L^{1/2}(U w_1 + e_u b_1)\|^2
  &= (U w_1 + e_u b_1)^{\top} L\,(U w_1 + e_u b_1) \nonumber \\
  &= \frac{1}{2}\sum_{i=1}^{p}\sum_{j=1}^{p} A_{ij}
     \bigl[(U_i w_1 + b_1)-(U_j w_1 + b_1)\bigr]^2
\label{eq:laplacian_expand}
\end{align}
 where $A_{ij}$ are the edge weights of the Universum graph.  This formulation
penalizes connected Universum samples whose projections onto the hyperplane
differ substantially, enforcing smoothness along the Universum manifold.
To solve Eq.~\eqref{eq:ug_obj1}, we define the augmented matrices:
 
\begin{align}
G &= [X_1 \ e_1]^T [X_1 \ e_1] + \delta I, \quad  & H &= [X_2 \ e_2]^T [X_2 \ e_2]\label{eq:GH}\\
P_{\text{graph}} &= [U\;e_u]^{\top} L\,[U\;e_u], &
z_1 &= [w_1;\,b_1] \label{eq:Pgraph}
\end{align}
Using these definitions, the numerator of Eq.~\eqref{eq:ug_obj1} becomes
$z_1^{\top} G z_1$ and the denominator becomes
$z_1^{\top}(H + P_{\text{graph}})z_1$, so the problem reduces to the Rayleigh
quotient minimization:
 
\begin{equation}
\min_{z_1 \neq 0}
  \frac{z_1^{\top} G z_1}{z_1^{\top}(H + P_{\text{graph}})z_1}
\label{eq:rayleigh1}
\end{equation}
Introducing the constraint $z_1^{\top}(H + P_{\text{graph}})z_1 = a_1$ for
some $a_1 > 0$ and forming the Lagrangian:
 
\begin{equation}
\mathcal{L}(z_1,\lambda_1)
  = z_1^{\top} G z_1
    - \lambda_1\!\left(z_1^{\top}(H + P_{\text{graph}})z_1 - a_1\right)
\label{eq:lagrangian1}
\end{equation}
Computing the gradient with respect to $z_1$ and setting it to zero yields:
 
\begin{equation}
\nabla_{z_1}\mathcal{L}
  = 2G z_1 - 2\lambda_1(H + P_{\text{graph}})z_1 = 0
\label{eq:gradient1}
\end{equation}
Rearranging gives the generalized eigenvalue problem:
 
\begin{equation}
G z_1 = \lambda_1\,(H + P_{\text{graph}})\,z_1
\label{eq:gep1}
\end{equation}
or equivalently:
 
\begin{equation}
\bigl([X_1\,e_1]^{\top}[X_1\,e_1] + \delta I\bigr)\,z_1
  = \lambda_1
    \bigl([X_2\,e_2]^{\top}[X_2\,e_2] + [U\,e_u]^{\top} L\,[U\,e_u]\bigr)\,z_1
\label{eq:gep1_explicit}
\end{equation}
The solution $z_1 = [w_1;\,b_1]^{\top}$ is the eigenvector associated with the smallest eigenvalue $\lambda_1$, providing hyperplane parameters that
jointly balance proximity to Class~1 samples, separation from Class~2 samples,
and smoothness along the Universum manifold.

For the second hyperplane, the analogous optimization problem is:
 
\begin{equation}
\min_{(w_2,\,b_2)\neq 0}
  \frac{\|X_2 w_2 + e_2 b_2\|^2 + \delta\,\|[w_2;\,b_2]\|^2}
       {\|X_1 w_2 + e_1 b_2\|^2 + \|L^{1/2}(U w_2 + e_u b_2)\|^2}
\label{eq:ug_obj2}
\end{equation}
Defining $M = [X_2\,e_2]^{\top}[X_2\,e_2] + \delta I$,
$N = [X_1\,e_1]^{\top}[X_1\,e_1]$,
$Q_{\text{graph}} = [U\,e_u]^{\top} L\,[U\,e_u]$, and $z_2 = [w_2;\,b_2]$,
the same Lagrangian procedure yields:
 
\begin{equation}
M z_2 = \lambda_2\,(N + Q_{\text{graph}})\,z_2
\label{eq:gep2}
\end{equation}
The solution $z_2 = [w_2;\,b_2]^{\top}$ is the eigenvector corresponding to
the smallest eigenvalue $\lambda_2$.
 
\subsection{Improved Universum Graph-Guided Generalized Eigenvalue Proximal
            Support Vector Machine (IUG-GEPSVM)}
 
IUG-GEPSVM extends the difference-based IU-GEPSVM formulation by replacing the
conventional Universum matrix $P = [U;\,e_u]^{\top}[U;\,e_u]$ with the
graph-regularized matrix $P_{\text{graph}} = [U;\,e_u]^{\top} L\,[U;\,e_u]$,
where $L$ is constructed from Universum samples via MST-based multi-hop
aggregation.  This preserves the numerical stability of IU-GEPSVM while
incorporating the geometric structure of the MCI manifold.  The objective
function for the first hyperplane is:
 
\begin{equation}
\min_{(w_1,\,b_1)\neq 0}
  \left(
    \frac{\|X_1 w_1 + e_1 b_1\|^2}{\|w_1\|^2 + b_1^2}
    - \gamma_1\,\frac{\|X_2 w_1 + e_2 b_1\|^2}{\|w_1\|^2 + b_1^2}
    - \psi_1\,\frac{\|L^{1/2}(U w_1 + e_u b_1)\|^2}{\|w_1\|^2 + b_1^2}
  \right)
\label{eq:iug_obj1}
\end{equation}
 where $\gamma_1 > 0$ controls class discrimination and $\psi_1 > 0$ controls the Universum contribution.  Using the matrix definitions from
Eqs.~\eqref{eq:GH}--\eqref{eq:Pgraph} and adding Tikhonov regularization,
Eq.~\eqref{eq:iug_obj1} is equivalently written as:
 
\begin{equation}
\min_{z_1 \neq 0}
  \left(
    \frac{z_1^{\top} G z_1}{z_1^{\top} z_1}
    - \gamma_1\,\frac{z_1^{\top} H z_1}{z_1^{\top} z_1}
    - \psi_1\,\frac{z_1^{\top} P_{\text{graph}} z_1}{z_1^{\top} z_1}
  \right)
\label{eq:iug_rayleigh}
\end{equation}
Introducing the constraint $z_1^{\top} z_1 = a_1$ and forming the Lagrangian:
 
\begin{equation}
\mathcal{L}(z_1,\lambda_1)
  = z_1^{\top}\!\left[(G+\delta I) - \gamma_1 H - \psi_1 P_{\text{graph}}\right]z_1
    - \lambda_1\!\left(z_1^{\top} z_1 - a_1\right)
\label{eq:iug_lagrangian}
\end{equation}
Setting the gradient to zero:
 
\begin{equation}
\nabla_{z_1}\mathcal{L}
  = 2\!\left[(G+\delta I) - \gamma_1 H - \psi_1 P_{\text{graph}}\right]z_1
    - 2\lambda_1 z_1 = 0
\label{eq:iug_gradient}
\end{equation}
yields the standard eigenvalue problem:
 
\begin{equation}
\left[(G + \delta I) - \gamma_1 H - \psi_1 P_{\text{graph}}\right] z_1
  = \lambda_1\, z_1
\label{eq:sep1}
\end{equation}
 
For the second hyperplane:
 
\begin{equation}
\left[(M + \delta I) - \gamma_2 N - \psi_2 Q_{\text{graph}}\right] z_2
  = \lambda_2\, z_2
\label{eq:sep2}
\end{equation}
where $M = [X_2\,e_2]^{\top}[X_2\,e_2] + \delta I$, $N = [X_1\,e_1]^{\top}[X_1\,e_1]$,
and $Q_{\text{graph}} = [U\,e_u]^{\top} L\,[U\,e_u]$.  The solutions $z_1$
and $z_2$ are the eigenvectors corresponding to the smallest eigenvalues.
By solving standard rather than generalized eigenvalue problems, IUG-GEPSVM
offers improved numerical stability over UG-GEPSVM, while the separate
weighting parameters $\gamma_i$ and $\psi_i$ provide independent control over
class separation and Universum alignment.
 
For classification, a new test sample $x$ is assigned to the class whose
hyperplane it lies closest to in terms of perpendicular distance:
 
\begin{equation}
\operatorname{class}(x)
  = \arg\min_{i\in\{1,2\}}\frac{|w_i^{\top} x + b_i|}{\|w_i\|}
\label{eq:classify}
\end{equation}

\section{Experiments and Results}
\label{experiments}

This section presents a comprehensive evaluation of the proposed UG-GEPSVM and
IUG-GEPSVM models against five established baselines on structural MRI extracted feature dataset
from the ADNI~\cite{mueller2005alzheimer} repository. The evaluation covers the classification Area Under the ROC Curve (AUC) under varying noise conditions, statistical significance testing, and hyperparameter sensitivity analysis.\\
\subsection{Experimental Setup and Dataset Description}
All experiments are conducted on an Intel Core i7-14700 processor (2.10~GHz)
with 16~GB RAM using Python~3.9.
Each dataset is split into 70\% training and 30\% testing samples.
Hyperparameters are selected by five-fold cross-validation on the training set.
The search grid covers $\{10^{-3},\allowbreak 10^{-2},\allowbreak
10^{-1},\allowbreak 10^{0},\allowbreak 10^{1},\allowbreak 10^{2}\}$ for all
scalar parameters and $\mathit{hops}\in\{0,1,2,3,4\}$ for the graph
propagation depth.
Classification performance is measured by the Area Under the ROC Curve (AUC) to account for the class imbalance between AD and CN subjects.

Structural MRI dataset are obtained from the UCSF Memory and Aging Center cohort
available in the Alzheimer's Disease Neuroimaging Initiative (ADNI)
repository~\cite{mueller2005alzheimer}, consisting of T1-weighted MRI scans.
The dataset contains 1,017 subjects: 788 cognitively normal (CN) subjects and
229 Alzheimer's disease (AD) subjects for AD versus CN classification.
Additionally, 391 mild cognitive impairment (MCI) subjects are used as
Universum data to provide intermediate structural information between the AD and
CN classes. Independent Component Analysis (ICA) and Principal Component Analysis (PCA) are applied for feature extraction, producing 155-dimensional
feature vectors for all subjects. To evaluate model robustness under realistic
acquisition variability, Gaussian noise is added at five levels (0\%, 5\%,
10\%, 15\%, and 20\%) to both ICA and PCA features, resulting in ten dataset
variants (five ICA-based and five PCA-based), each of size $1017 \times 155$.

\subsection{Results Analysis}
\label{sec:results}

Tables~\ref{tab:ica} and~\ref{tab:pca} report AUC values on the ICA-based and PCA-based dataset variants for all seven classifiers together with their best hyperparameter configurations, respectively.

\begin{table*}[htbp]
\caption{Performance comparison (AUC\,$\uparrow$) on ICA-based dataset
variants.}
\label{tab:ica}
\centering
\setlength{\tabcolsep}{4pt}
\renewcommand{\arraystretch}{1.8}
\hspace*{-0.26\textwidth}\resizebox{1.5\textwidth}{!}{%
\begin{tabular}{|c|c|c|c|c|c|c|c|}
\hline
 &
\textbf{GEPSVM}~\cite{mangasarian2006multisurface} &
\textbf{IGEPSVM}~\cite{shao2012improved} &
\textbf{UTSVM}~\cite{qi2012twin} &
\textbf{UGEPSVM}~\cite{kumar2026unified} &
\textbf{IUGEPSVM}~\cite{kumar2026unified} &
\textbf{IUG-GEPSVM}$^*$ &
\textbf{UG-GEPSVM}$^*$ \\
\hline
\textbf{Dataset} &
\textbf{AUC\,$\uparrow$} & \textbf{AUC\,$\uparrow$} &
\textbf{AUC\,$\uparrow$} & \textbf{AUC\,$\uparrow$} &
\textbf{AUC\,$\uparrow$} & \textbf{AUC\,$\uparrow$} &
\textbf{AUC\,$\uparrow$} \\
$(Patterns\times Features)$ &
$(\delta,\gamma)$ &
$(\delta,\nu,\gamma)$ &
$(c,c_u,\epsilon,\mathrm{reg},\gamma)$ &
$(\delta,\gamma)$ &
$(\delta,\gamma,\psi,\sigma)$ &
$(\delta,\gamma_1,\gamma_2,\psi_1,\psi_2,\mathit{hops})$ &
$(\delta,\mathit{hops},\gamma)$ \\
\hline
0\%-ICA &
85.50 & 83.80 & 71.90 & 86.20 & 86.49 & 87.07 & \textbf{88.40} \\
$(1017{\times}155)$ &
$(10^{-3},10^{-2})$ &
$(10^{-3},10^{-3},10^{-2})$ &
$(10^{-1},10^{0},10^{-1},10^{0},10^{-1})$ &
$(10^{-3},10^{2})$ &
$(10^{-1},10^{-1},10^{-3},10^{1})$ &
$(10^{-3},10^{-3},10^{-3},10^{-3},10^{-3},0)$ &
$(10^{-3},1,10^{2})$ \\
\hline
5\%-ICA &
85.50 & 84.00 & 72.20 & 86.20 & 86.66 & 86.76 & \textbf{88.30} \\
$(1017{\times}155)$ &
$(10^{-3},10^{-2})$ &
$(10^{-3},10^{-3},10^{-2})$ &
$(10^{-1},10^{0},10^{-1},10^{0},10^{-1})$ &
$(10^{-3},10^{2})$ &
$(10^{-1},10^{-1},10^{-3},10^{1})$ &
$(10^{-3},10^{-3},10^{-3},10^{-3},10^{-3},4)$ &
$(10^{-3},2,10^{2})$ \\
\hline
10\%-ICA &
71.70 & 84.20 & 72.00 & 87.00 & 86.78 & 87.42 & \textbf{89.70} \\
$(1017{\times}155)$ &
$(10^{-3},10^{-2})$ &
$(10^{-3},10^{-3},10^{-2})$ &
$(10^{-1},10^{0},10^{-1},10^{0},10^{-1})$ &
$(10^{-3},10^{2})$ &
$(10^{-1},10^{-1},10^{-3},10^{1})$ &
$(10^{-3},10^{-3},10^{-3},10^{-3},10^{-3},4)$ &
$(10^{-3},0,10^{2})$ \\
\hline
15\%-ICA &
72.90 & 84.00 & 72.90 & 84.00 & 86.49 & 86.80 & \textbf{88.00} \\
$(1017{\times}155)$ &
$(10^{-3},10^{-2})$ &
$(10^{-3},10^{-3},10^{-2})$ &
$(10^{-1},10^{0},10^{-1},10^{0},10^{-1})$ &
$(10^{-3},10^{2})$ &
$(10^{-1},10^{-1},10^{-3},10^{1})$ &
$(10^{-3},10^{-3},10^{-3},10^{-3},10^{-3},1)$ &
$(10^{-3},2,10^{2})$ \\
\hline
20\%-ICA &
69.20 & 82.20 & 69.40 & 86.60 & 85.96 & 85.46 & \textbf{88.00} \\
$(1017{\times}155)$ &
$(10^{-3},10^{-1})$ &
$(10^{-3},10^{-3},10^{-2})$ &
$(10^{-1},10^{0},10^{-1},10^{0},10^{-1})$ &
$(10^{-3},10^{2})$ &
$(10^{-1},10^{-1},10^{-3},10^{1})$ &
$(10^{-3},10^{-3},10^{-3},10^{-3},10^{-3},0)$ &
$(10^{-3},1,10^{2})$ \\
\hline
\textbf{Average AUC\,$\uparrow$} &
76.96 & 83.64 & 71.68 & 86.00 & 86.48 & 86.70 & \textbf{88.48} \\
\hline
\textbf{Average Rank\,$\downarrow$} &
6.10 & 5.30 & 6.50 & 3.50 & 3.20 & 2.40 & \textbf{1.00} \\
\hline
\multicolumn{8}{l}{$^*$ denotes the proposed model.}
\end{tabular}}
\end{table*}
Average AUC and average rank (lower is better) are
reported at the bottom of each table. The average rank is computed by ranking
all seven methods on each dataset from best (rank 1) to worst (rank 7) and
averaging over all ten variants.
\begin{table*}[htbp]
\caption{Performance comparison (AUC\,$\uparrow$) on PCA-based dataset
variants. }
\label{tab:pca}
\centering
\setlength{\tabcolsep}{4pt}
\renewcommand{\arraystretch}{1.8}
\hspace*{-0.26\textwidth}\resizebox{1.5\textwidth}{!}{%
\begin{tabular}{|c|c|c|c|c|c|c|c|}
\hline
 &
\textbf{GEPSVM}~\cite{mangasarian2006multisurface} &
\textbf{IGEPSVM}~\cite{shao2012improved} &
\textbf{UTSVM}~\cite{qi2012twin} &
\textbf{UGEPSVM}~\cite{kumar2026unified} &
\textbf{IUGEPSVM}~\cite{kumar2026unified} &
\textbf{IUG-GEPSVM}$^*$ &
\textbf{UG-GEPSVM}$^*$ \\
\hline
\textbf{Dataset} &
\textbf{AUC\,$\uparrow$} & \textbf{AUC\,$\uparrow$} &
\textbf{AUC\,$\uparrow$} & \textbf{AUC\,$\uparrow$} &
\textbf{AUC\,$\uparrow$} & \textbf{AUC\,$\uparrow$} &
\textbf{AUC\,$\uparrow$} \\
$(Patterns\times Features)$ &
$(\delta,\gamma)$ &
$(\delta,\nu,\gamma)$ &
$(c,c_u,\epsilon,\mathrm{reg},\gamma)$ &
$(\delta,\gamma)$ &
$(\delta,\gamma,\psi,\sigma)$ &
$(\delta,\gamma_1,\gamma_2,\psi_1,\psi_2,\mathit{hops})$ &
$(\delta,\mathit{hops},\gamma)$ \\
\hline
0\%-PCA &
85.70 & 83.90 & 72.40 & 86.30 & 86.42 & 86.50 & \textbf{87.00} \\
$(1017{\times}155)$ &
$(10^{-3},10^{-1})$ &
$(10^{-3},10^{-3},10^{-2})$ &
$(10^{-1},10^{0},10^{-1},10^{0},10^{-1})$ &
$(10^{-3},10^{2})$ &
$(10^{-1},10^{-1},10^{-3},10^{1})$ &
$(10^{-3},10^{-3},10^{-3},10^{-3},10^{-3},2)$ &
$(10^{-3},1,10^{2})$ \\
\hline
5\%-PCA &
85.80 & 84.20 & 72.40 & 86.40 & 86.75 & 86.62 & \textbf{87.40} \\
$(1017{\times}155)$ &
$(10^{-3},10^{-1})$ &
$(10^{-3},10^{-3},10^{-2})$ &
$(10^{-1},10^{0},10^{-1},10^{0},10^{-1})$ &
$(10^{-3},10^{2})$ &
$(10^{-1},10^{-1},10^{-3},10^{1})$ &
$(10^{-3},10^{-3},10^{-3},10^{-3},10^{-3},0)$ &
$(10^{-3},1,10^{2})$ \\
\hline
10\%-PCA &
72.50 & 83.90 & 72.40 & 87.80 & 86.50 & 87.57 & \textbf{88.10} \\
$(1017{\times}155)$ &
$(10^{-3},10^{-1})$ &
$(10^{-3},10^{-3},10^{-2})$ &
$(10^{-1},10^{0},10^{-1},10^{0},10^{-1})$ &
$(10^{-3},10^{2})$ &
$(10^{-1},10^{-1},10^{-3},10^{1})$ &
$(10^{-3},10^{-3},10^{-3},10^{-3},10^{-3},0)$ &
$(10^{-3},4,10^{2})$ \\
\hline
15\%-PCA &
71.70 & 84.20 & 71.60 & 87.00 & 87.17 & 85.62 & \textbf{87.50} \\
$(1017{\times}155)$ &
$(10^{-3},10^{-1})$ &
$(10^{-3},10^{-2},10^{-2})$ &
$(10^{-1},10^{0},10^{-1},10^{0},10^{-1})$ &
$(10^{-3},10^{2})$ &
$(10^{-1},10^{-1},10^{-3},10^{1})$ &
$(10^{-3},10^{-3},10^{-3},10^{-3},10^{-3},0)$ &
$(10^{-3},1,10^{2})$ \\
\hline
20\%-PCA &
73.20 & 83.10 & 72.70 & 88.20 & 86.68 & 86.42 & \textbf{88.30} \\
$(1017{\times}155)$ &
$(10^{-3},10^{-1})$ &
$(10^{-3},10^{-2},10^{-2})$ &
$(10^{-1},10^{0},10^{-1},10^{0},10^{-1})$ &
$(10^{-3},10^{2})$ &
$(10^{-1},10^{-1},10^{-3},10^{1})$ &
$(10^{-3},10^{-3},10^{-3},10^{-3},10^{-3},4)$ &
$(10^{-3},2,10^{2})$ \\
\hline
\textbf{Average AUC\,$\uparrow$} &
77.78 & 83.86 & 72.30 & 87.14 & 86.70 & 86.55 & \textbf{87.66} \\
\hline
\textbf{Average Rank\,$\downarrow$} &
5.60 & 5.40 & 7.00 & 3.00 & 2.80 & 3.20 & \textbf{1.00} \\
\hline
\multicolumn{8}{l}{$^*$ denotes the proposed model.}
\end{tabular}}
\end{table*}
Tables~\ref{tab:ica} and~\ref{tab:pca} compare GEPSVM, IGEPSVM, UTSVM, UGEPSVM, \\IUGEPSVM, and the proposed IUG-GEPSVM and UG-GEPSVM models on 10 AD versus CN dataset variants generated using ICA and PCA features at five noise levels (0\%--20\%). Across the ICA datasets, IUGEPSVM and UGEPSVM achieve average AUC values of 86.48\% and 86.00\%, respectively, while on the PCA datasets their average AUC values are 86.70\% and 87.14\%. Considering all ten dataset variants together, IUGEPSVM and UGEPSVM obtain the highest average baseline AUC values of 86.59\% and 86.57\%, respectively, followed by IGEPSVM (83.75\%), GEPSVM (77.37\%), and UTSVM (71.99\%). These results indicate that incorporating Universum information consistently improves classification performance over conventional GEPSVM-based models that do not exploit prior structural knowledge.

\begin{figure}
  \centering
  \includegraphics[width=0.8\columnwidth]{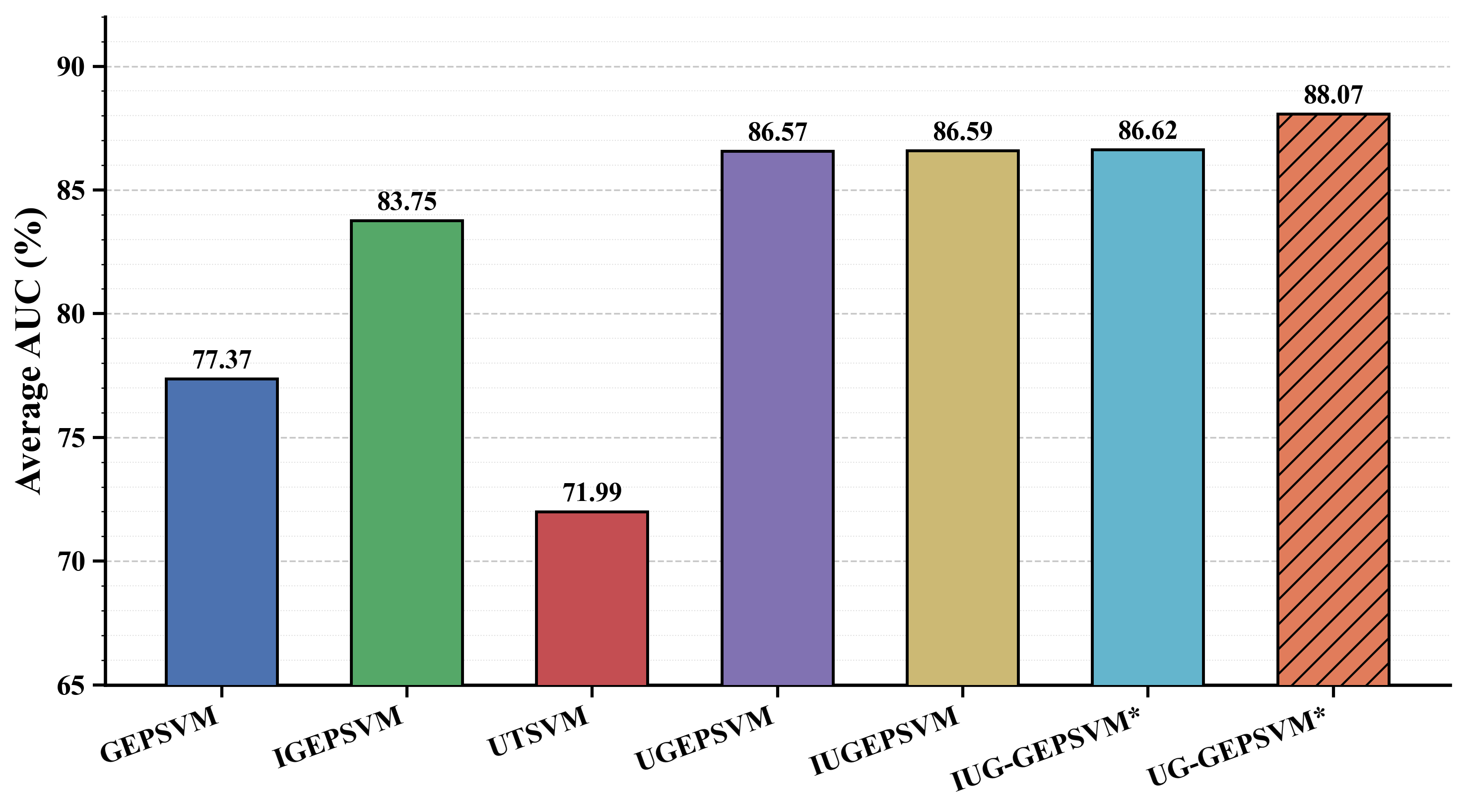}
  \caption{Average AUC (\%) across all dataset variants for each compared method. Proposed models are marked with $^*$.}
  \label{fig:bar}
\end{figure}

The proposed UG-GEPSVM achieves the highest average AUC of 88.07\% across all ten variants and the best average rank of 1.00. Its peak AUC of 89.70\% is obtained on the 10\%-ICA dataset, the highest single result among all compared methods. IUG-GEPSVM achieves the second-best average AUC of 86.62\%. Compared with UGEPSVM and IGEPSVM, UG-GEPSVM improves the average AUC by
$+1.48\%$ and $+4.32\%$, respectively.
These improvements demonstrate that incorporating graph-structural information
from MCI Universum samples provides additional discriminative signal beyond
conventional Universum regularization, which treats each MCI sample in
isolation.
Figure~\ref{fig:bar} further visualises the overall average AUC for each
method, making the consistent superiority of UG-GEPSVM immediately apparent
across both feature representations.

\noindent

The advantage of the proposed graph-guided models becomes particularly evident as the noise level increases. On the ICA datasets, the AUC of GEPSVM decreases sharply from 85.50\% at 0\% noise to 69.20\% at 20\% noise, corresponding to a degradation of 16.30 percentage points. UTSVM exhibits a similar decline, dropping from 71.90\% to 69.40\%. In contrast, UG-GEPSVM maintains AUC values between 88.00\% and 89.70\% across all ICA noise levels, resulting in a variation of only 1.70 percentage points. A similar trend is observed on the PCA datasets, where the AUC of UG-GEPSVM remains within a narrow range of 87.00\% to 88.30\% regardless of the noise level. Although IGEPSVM, UGEPSVM, and IUGEPSVM demonstrate reasonable robustness by consistently achieving AUC values above 82.20\%, none of these methods matches the stability and consistency of UG-GEPSVM.

\begin{figure}[htbp]
  \centering
  \includegraphics[width=0.95\columnwidth]{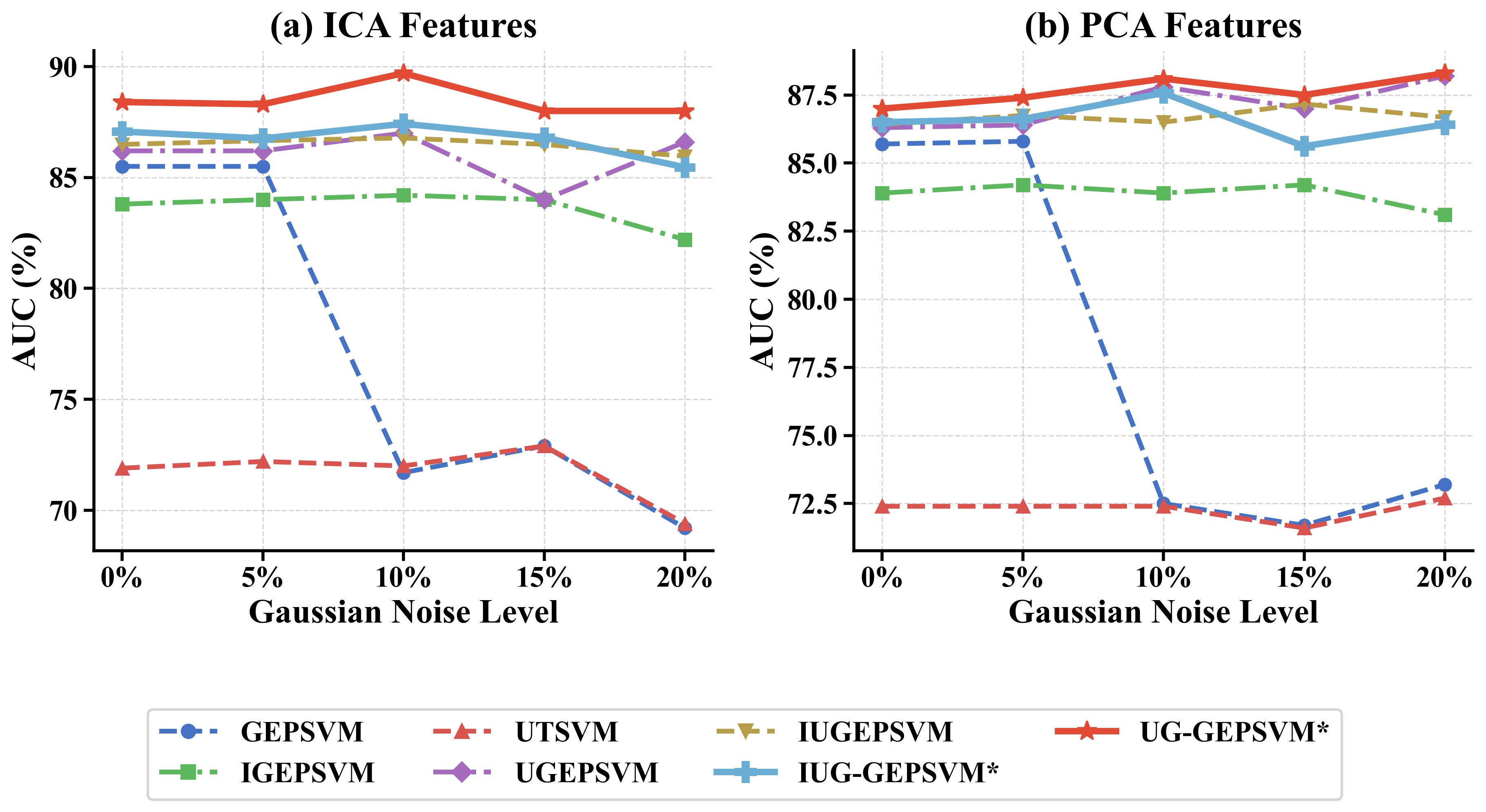}
  \caption{AUC (\%) versus Gaussian noise level for ICA features (left panel) and PCA features (right panel). Proposed UG-GEPSVM$^*$ maintains the highest and most stable AUC across all noise levels on both feature representations.}
  \label{fig:noise}
\end{figure}
The robustness of UG-GEPSVM can be explained from a geometric perspective. Isotropic Gaussian noise perturbs feature directions in a nearly uniform manner throughout the feature space. The MST-based graph Laplacian regularization constrains the decision hyperplanes to vary smoothly along the underlying MCI manifold, which forms a compact and locally connected structure within the 155-dimensional ICA and PCA feature spaces. This manifold is constructed using pairwise Gaussian similarities and MST connectivity, both of which remain relatively stable under small isotropic perturbations. As a result, the graph-regularized hyperplanes are less sensitive to noise than hyperplanes learned solely from raw Euclidean distances. Furthermore, multi-hop aggregation propagates information across neighbouring nodes, reducing the influence of local noise artefacts while preserving the global manifold structure. Together, these properties allows UG-GEPSVM to maintain high and stable AUC values even at 20\% noise, a setting in which conventional Euclidean-distance-based methods experience substantial performance degradation. Figure~\ref{fig:noise} plots AUC against noise level for all seven methods on
ICA and PCA features simultaneously, providing a clear visual confirmation of
the robustness advantage.

\subsection{Statistical Analysis}
\label{sec:stats}

This subsection evaluates the statistical significance of the performance differences among the compared classifiers.

\begin{figure}[htbp]
  \centering
  \includegraphics[width=0.8\columnwidth]{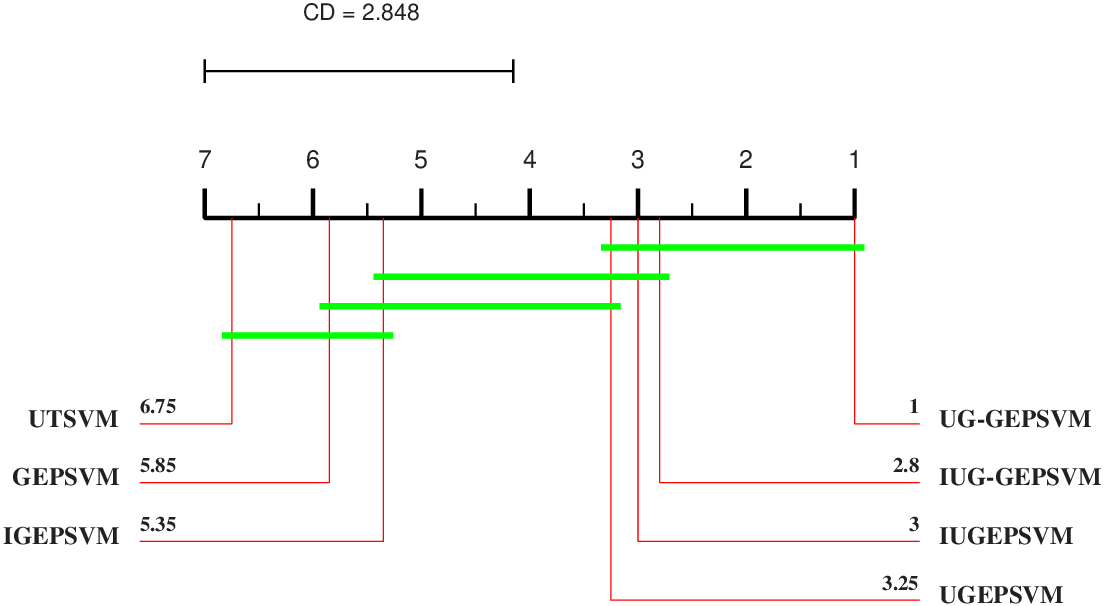}
  \caption{Critical difference diagram comparing average AUC ranks of the
           baseline methods and the proposed IUG-GEPSVM and UG-GEPSVM.
           Methods connected by a horizontal bar are not significantly
           different at the $5\%$ level ($\mathrm{CD}=2.85$).}
  \label{fig:cd_diagram}
\end{figure}

\noindent\textbf{Friedman test.}
The Friedman test is used to compare the performance of seven classifiers on
the 10 AD versus CN datasets.
Based on the AUC values, the average ranks of GEPSVM, IGEPSVM, UTSVM,
UGEPSVM, IUGEPSVM, IUG-GEPSVM, and UG-GEPSVM are 5.85, 5.35, 6.75, 3.25, 3.00, 2.80, and 1.00, respectively (computed over all ten ICA- and PCA-based variants jointly).
The proposed UG-GEPSVM achieves the best (lowest) rank, followed by
IUG-GEPSVM.
For $N=10$ datasets and $K=7$ classifiers, the Friedman statistic is
$\chi_F^2 = 53.1643$ and the corresponding Iman--Davenport statistic is
$F_F = 69.9970$.
Since the obtained $F_F$ value exceeds the critical value at the $5\%$
significance level, the null hypothesis of equal classifier performance is
rejected, confirming that statistically significant differences exist among the
compared methods.

\noindent\textbf{Nemenyi post-hoc test.}
The Nemenyi post-hoc test is applied to identify which pairs of classifiers
differ significantly.
The critical difference (CD) at the $5\%$ significance level is $\mathrm{CD}
= 2.85$.
As shown in Fig.~\ref{fig:cd_diagram}, the proposed IUG-GEPSVM and UG-GEPSVM
models achieve the two best average ranks, with UG-GEPSVM obtaining the
overall best performance.
Classifiers whose average ranks differ by more than the CD are considered
significantly different; both proposed models are significantly better than
GEPSVM, IGEPSVM, and UTSVM.
These results confirm the effectiveness of the proposed graph-guided Universum
learning framework for Alzheimer's disease detection.

\subsection{Sensitivity Analysis of Hyperparameters}
\label{sec:sensitivity}

\begin{figure}[htbp]
  \centering
  \begin{minipage}[b]{0.43\textwidth}
    \centering
    \includegraphics[width=0.9\textwidth]{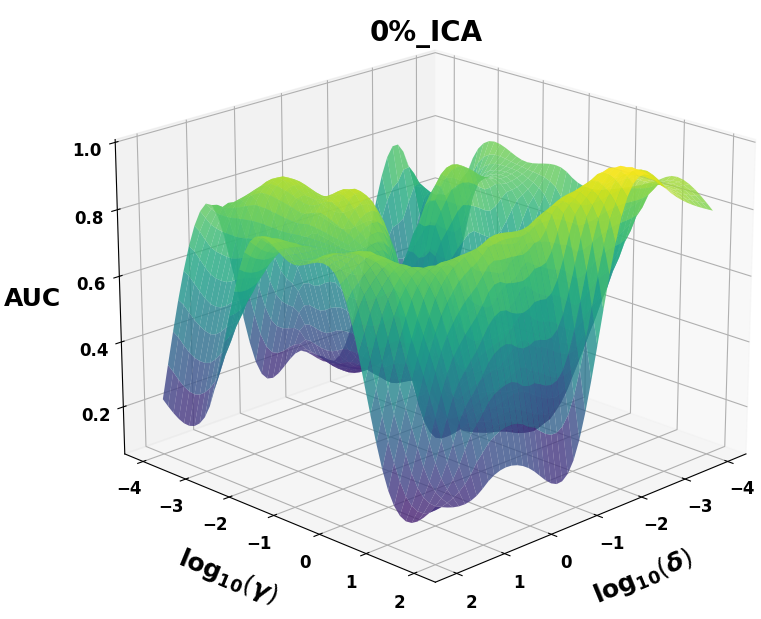}
  \end{minipage}\hfill
  \begin{minipage}[b]{0.43\textwidth}
    \centering
    \includegraphics[width=0.9\textwidth]{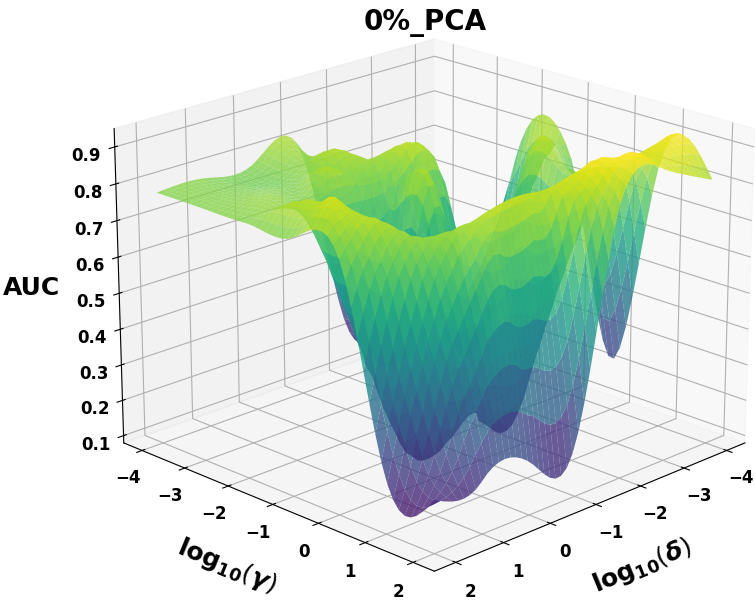}
  \end{minipage}\\[0.80ex]
  \begin{minipage}[b]{0.43\textwidth}
    \centering
    \includegraphics[width=0.9\textwidth]{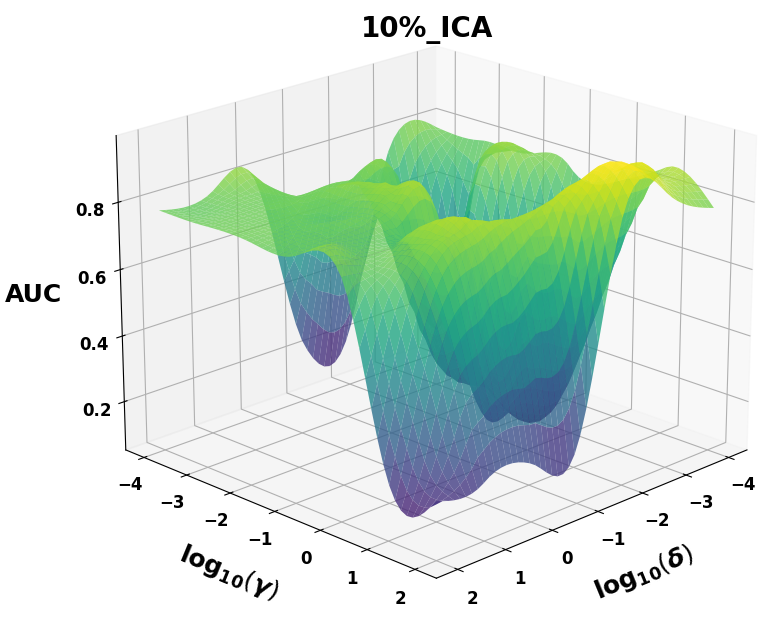}
  \end{minipage}\hfill
  \begin{minipage}[b]{0.43\textwidth}
    \centering
    \includegraphics[width=0.9\textwidth]{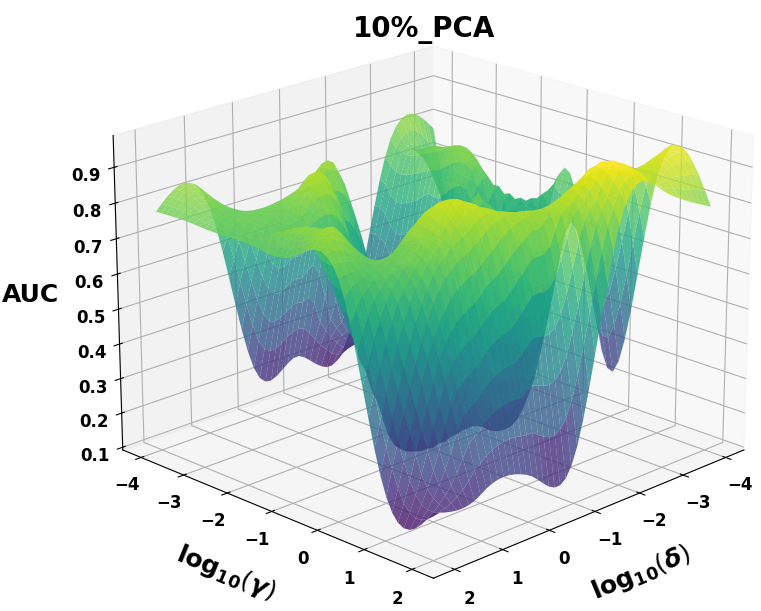}
  \end{minipage}
  \caption{AUC variation with $\gamma$ and $\delta$ for the proposed model
           UG-GEPSVM on (top-left) 0\%-ICA, (top-right) 0\%-PCA,
           (bottom-left) 10\%-ICA, and (bottom-right) 10\%-PCA variants.}
  \label{fig:sensitivity}
\end{figure}

Figure~\ref{fig:sensitivity} shows the variation of AUC with respect to
$\log_{10}(\gamma)$ and $\log_{10}(\delta)$ for UG-GEPSVM on four
representative dataset variants: 0\%-ICA, 0\%-PCA, 10\%-ICA, and 10\%-PCA. Across all four datasets with different noise levels, UG-GEPSVM demonstrates stable performance over a wide
range of hyperparameter values. The highest AUC values are generally observed
in the central region of the parameter space, while performance decreases when
either parameter approaches its extreme values.

\noindent For the ICA-based datasets (0\%-ICA and 10\%-ICA), the AUC reaches values
close to 0.90 within the region $\log_{10}(\gamma) \in [-2,0]$ and
$\log_{10}(\delta) \in [-2,0]$.
Similar behaviour is observed for the PCA-based datasets (0\%-PCA and
10\%-PCA), where the highest AUC values during training approach 0.88.
In both feature representations, a broad high-performance plateau is observed around moderate parameter values. The sensitivity surfaces remain largely similar under both 0\% and 10\% noise conditions. This observation indicates that the performance of UG-GEPSVM is relatively stable with respect to hyperparameter selection, even in the presence of noise. The wide plateau of high AUC values suggests that the model does not require highly precise parameter tuning and can achieve competitive performance over a broad range of $\gamma$ and $\delta$ values.

\section{Conclusion}
\label{conclusion}

This paper proposed two graph-guided Universum learning models, UG-GEPSVM and IUG-GEPSVM, for AD versus CN classification using structural MRI data. The proposed framework incorporates a graph Laplacian regularization term constructed from MCI Universum samples using Gaussian similarity, MST connectivity, and multi-hop aggregation. By exploiting both prior Universum information and the underlying geometric structure of the MCI manifold, the proposed models encourage smoother decision boundaries and improved generalization. Experimental evaluations on ten ICA- and PCA-based dataset variants under different Gaussian noise levels demonstrated the effectiveness of the proposed approach. Among all compared methods, UG-GEPSVM achieved the best overall performance with an average AUC of 88.07\% across dataset variants. The model also exhibited strong robustness to noise, maintaining stable performance even at higher noise levels. Statistical analyses, including the Friedman and Nemenyi tests, further confirmed the significance of the observed improvements over existing GEPSVM- and Universum-based methods. Future work will focus on adaptive graph construction strategies, kernelized graph-guided formulations for nonlinear learning, and extensions to multi-class Alzheimer's disease classification. Integrating multimodal biomarkers, such as PET imaging within the proposed graph-guided Universum framework also represents a promising direction for future research.

\bibliographystyle{elsarticle-num}
\bibliography{bibliography}
\nocite{*}
\end{document}